\documentclass[sigconf]{acmart}

\usepackage{algorithm}
\usepackage{algorithmic}
\usepackage{subfigure}
\usepackage{graphicx}

\AtBeginDocument{%
  }

\setcopyright{rightsretained}
\copyrightyear{2024}
\acmDOI{}

\acmConference[KDD '24]{3nd Workshop on End-End Customer Journey Optimization at KDD 2024.}{August 25--29,2024}{Barcelona, Spain}
\acmISBN{}

\begin{document}

\title{Towards Lifelong Learning Embeddings: An Algorithmic Approach to Dynamically Extend Embeddings}

\author{Miguel Alves Gomes}
\orcid{0000-0003-3664-0360}
\affiliation{%
  \institution{Institute for Technologies and Management of Digital Transformation, University of Wuppertal}
  \streetaddress{Rainer-Gruenter-Str. 21}
  \city{Wuppertal}
  \state{North Rhine-Westphalia}
  \country{Germany}
}
\email{alvesgomes@uni-wuppertal.de}

\author{Philipp Meisen}
\orcid{0000-0002-8024-3074}
\affiliation{
  \institution{Breinify Inc.}
  \city{San Francisco}
  \state{California}
  \country{USA}
}
\email{philipp.meisen@breinify.com}

\author{Tobias Meisen}
\orcid{0000-0002-1969-559X}
\affiliation{%
  \institution{Institute for Technologies and Management of Digital Transformation, University of Wuppertal}
  \streetaddress{Rainer-Gruenter-Str. 21}
  \city{Wuppertal}
  \state{North Rhine-Westphalia}
  \country{Germany}
}
\email{meisen@uni-wuppertal.de}


\begin{abstract}
The rapid evolution of technology has transformed business operations and customer interactions worldwide, with personalization emerging as a key opportunity for e-commerce companies to engage customers more effectively. The application of machine learning, particularly that of deep learning models, has gained significant traction due to its ability to rapidly recognize patterns in large datasets, thereby offering numerous possibilities for personalization. These models use embeddings to map discrete information, such as product IDs, into a latent vector space, a method increasingly popular in recent years. However, e-commerce's dynamic nature, characterized by frequent new product introductions, poses challenges for these embeddings, which typically require fixed dimensions and inputs, leading to the need for periodic retraining from scratch. This paper introduces a modular algorithm that extends embedding input size while preserving learned knowledge, addressing the challenges posed by e-commerce's dynamism. The proposed algorithm also incorporates strategies to mitigate the cold start problem associated with new products. The results of initial experiments suggest that this method outperforms traditional embeddings.
\end{abstract}




\maketitle

\section{Introduction}
The advent of new technologies and devices has led to a remarkable transformation of businesses and customer interactions around the globe in recent years~\cite{CecileFerrera.2019}. In this context, personalization offers new opportunities for e-commerce companies by enabling targeted and individualized customer engagement~\cite{MiguelAlvesGomes.2023.review}. Nevertheless, to achieve this, a large amount of interaction data needs to be collected, which in turn allows search engines and recommendation systems to be continuously improved~\cite{ZhifangFan.2022, GuoruiZhou.2019,DavidCarmel.2020}, through the ability to utilize the potential of machine learning. Thereby, machine learning has gained immense popularity in general due to its ability to rapidly recognize patterns in vast amounts of data, swiftly visualize analytics, and provide in-depth insights~\cite{NehaSoni.2020,DanielaFreddi.2018,IainCockburn.2018}. Especially, with the ongoing development of deep learning models like deep-attention-based networks for predicting customer click behavior like DIEN~\cite{GuoruiZhou.2019} or transformer-based models like Bert4Rec~\cite{FeiSun.2019} for product recommendations are gaining huge popularity. To extract information from the available data and represent customer behavior, these deep learning approaches use embeddings to project discrete information, such as product IDs, into a latent vector space, as shown by many researches in the last years~\cite{MiguelAlvesGomes.2022,MiguelAlvesGomes.2023,Sheil.2018,MiguelAlvesGomes.2021,Tercan.2021,Srilakshmi.2022,Vasile.2016,JiangweiZeng.2020,GuojingHuang.2020,ChiChen.2022,ChangZhou.2018}. 

\begin{figure}[htbp]
    \centering
    \subfigure{
        \includegraphics[width=0.45\textwidth]{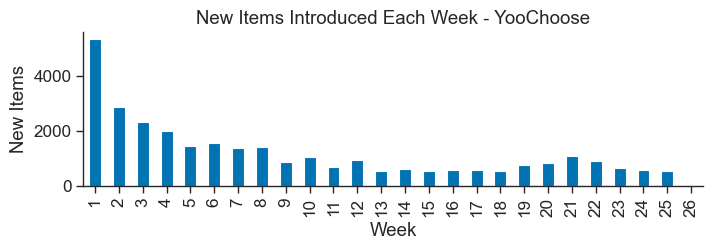}
        \label{fig:subfig1}
    }
    \subfigure{
        \includegraphics[width=0.45\textwidth]{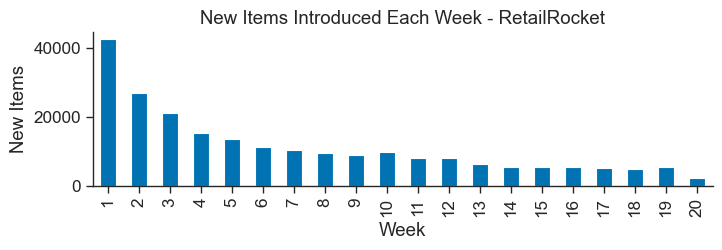}
        \label{fig:subfig2}
    }
    \caption{Number of new items per week for the YooChoose and RetailRocket benchmark datasets.}
    \label{fig:main}
\end{figure}

E-commerce is a highly competitive and dynamic field, characterized by the rapid introduction of new products and services. Figure \ref{fig:main} illustrates the weekly introduction of new products in two prominent benchmark datasets: YooChoose, for customer purchase prediction, and RetailRocket, for customer churn prediction. However, for embeddings such as those used in aforementioned deep learning models, this dynamic is a problem. These embeddings typically employ fixed dimensions and fixed inputs, necessitating periodic retraining from scratch to account for new products and evolving trends.

To address this problem, we introduce a streamlined and modular algorithm designed to extend the embedding input size while preserving previously acquired knowledge. This approach effectively addresses the aforementioned problem. Furthermore, the algorithm supports the implementation of various strategies to mitigate the cold start problem associated with newly introduced products. Preliminary experiments demonstrate that our method yields performance improvements over traditional input embeddings.

The structure of our paper is as follows: Section 2 presents a comprehensive review of related work, emphasizing approaches that address the aforementioned problem in natural language processing (NLP). In this section, we also analyze the limitations of these methods when applied to the embeddings commonly used in e-commerce. Section 3 details our proposed algorithm, providing a thorough explanation of its design and implementation. In Section 4, we present the results of our preliminary experiments and discuss their implications. Finally, Section 5 concludes the paper with a discussion on future research directions and outstanding questions that need to be addressed to validate the effectiveness of our proposed algorithm.

\section{Related Work and Its Limitations for E-Commerce}
Advancements and achievements in Natural Language Processing (NLP) leveraging foundation models have been significantly driven by key technologies such as self-attention mechanisms~\cite{AshishVaswani.2017} and word embeddings, as introduced by Mikolov et al.~\cite{TomasMikolov.2013} in 2013. Thereby, the "Out-of-Vocabulary" (OoV) problem posed a substantial challenge in the language domain, a challenge that we seek to address in the context of embeddings used in e-commerce. A straightforward solution to the OoV problem is the introduction of an unknown token~\cite{IlyaSutskever.2014,MiguelAlvesGomes.2022}, a technique that has been prevalent in NLP and continues to be a popular method for managing newly introduced tokens in e-commerce.

However, unlike the relatively static nature of language, which evolves over years, the e-commerce domain is highly dynamic. For instance, the continuous influx of new products results in a predominance of unknown tokens, often surpassing the number of recognized tokens. This dynamic renders the unknown token approach less effective, especially in environments where new products are frequently introduced. While this method may suffice in scenarios with infrequent updates, such as in language, it is inadequate for addressing the rapidly changing landscape of e-commerce. 

Another widely used method in NLP is subword embeddings~\cite{RicoSennrich.2015,AlecRadford.2019,YinhanLiu.2019}, which can be generated using techniques such as byte-pair encoding~\cite{AshishVaswani.2017}. This approach involves splitting words into their constituent subwords, allowing each word to be reconstructed from these smaller units. Although effective in NLP, this method is not applicable in the e-commerce domain. Unlike words, tokens representing products cannot be decomposed and reassembled from other products, thereby limiting the utility of subword embeddings in this context.

To the best of our knowledge, there is currently no suitable method in the e-commerce sector for integrating new tokens, such as products, into embeddings. Therefore, it is essential to develop extensible embeddings that can adapt to the dynamic nature of e-commerce, ensuring their applicability in a productive environment. The algorithmic approach presented here represents a preliminary attempt to address the specific demands of an e-commerce environment.

\section{Algorithm}
Embeddings are typically implemented as a single-layer neural network that projects a token into an n-dimensional vector space, incorporating the context of the token during training. In the context of e-commerce, it is either used as pretrained embeddings on customer interactions and used as features for a learning model or recommendation~\cite{MiguelAlvesGomes.2022,MiguelAlvesGomes.2023,Vasile.2016,Tercan.2021} or directly in an end-to-end architecture that learns representation and task solving or recommendation at once~\cite{JiangweiZeng.2020,GuojingHuang.2020,ChiChen.2022,ChangZhou.2018}. In typical applications, the input to the embedding layer is generally presented in the form of one-hot encoding. Consequently, the embedding layer effectively functions as a collection of weights corresponding to each input token. This design permits the interchangeability of tokens by substituting their associated weights without any loss of information. Furthermore, it is possible to expand the input space by incorporating additional weights, provided that the embedding dimension remains constant. The most straightforward approach to this expansion involves initializing random weights for the new tokens. However, this approach can result in suboptimal performance, a phenomenon commonly referred to as the cold-start problem~\cite{BlerinaLika.2014}.

To address this issue, various strategies can be employed for selecting new weights. Examples of such strategies for products include:
\begin{enumerate}
    \item Using the weights of the unknown token,
    \item Computing the average weights of all tokens,
    \item Computing the average weights of all product tokens within the same product category,
    \item Take the weights of a product with similar features. 
\end{enumerate}

These methods aim to facilitate a more informed initialization of weights, thereby addressing the cold-start problem and improving the performance of embeddings in dynamic e-commerce environments. Algorithm~\ref{alg:update_embedding} shows our proposed approach to extend the input size of the embedding\footnote{Algorithm will be published over github in camera-ready version}. As the algorithm iterates over all tokens in the new map, it facilitates both the extension and reduction of tokens within the embedding, since tokens that are absent from the new map are not transferred.

\begin{algorithm}
\caption{Update Embedding Weights for New Tokens}
\label{alg:update_embedding}
\begin{algorithmic}[1]
\REQUIRE 
\STATE new\_map: mapping new tokens to ids
\STATE old\_map: mapping old tokens to ids
\STATE new\_embedding: new generated embedding with the input size = num new tokens and dim N
\STATE old\_embedding: pretrained embeddings with weights and dim N
\STATE new\_weights: strategy to apply new weights to the embedding
\ENSURE Updates the embedding weights for the new tokens

\FOR{ \textbf{each} (token, id) \textbf{in} new\_map }
    \IF{ token \textbf{in} old\_map }
        \STATE new\_embedding[id] $\gets$ old\_embedding[old\_map[token]]
    \ELSE
        \STATE new\_embedding[id] $\gets$ new\_weights
    \ENDIF
\ENDFOR

\end{algorithmic}
\end{algorithm}

\section{Initial Experiments and Results}
To evaluate the effectiveness of incremental learning embeddings and ascertain their feasibility, we conducted a series of preliminary experiments. Our focus was on the purchase prediction use case, utilizing the widely recognized yoochoose benchmark dataset\footnote{\url{https://www.kaggle.com/datasets/chadgostopp/recsys-challenge-2015} (accessed 2024-06-01)}. The objective of this benchmark is to predict whether a purchase will occur based on users' product viewing history within a session.

To assess the incremental learning algorithm, we partitioned the dataset into 26 segments, each corresponding to one week. The model was trained using data from one week and its performance was evaluated on the subsequent week's data. For these initial experiments, we implemented a straightforward end-to-end embedding LSTM architecture that processes a sequence of products and performs binary classification to predict the likelihood of a purchase.

We investigated four different approaches, as described below:
\begin{enumerate}
    \item \textbf{Baseline}: No incremental learning; each week is learned from scratch.
    \item \textbf{Random}: Incremental learning with new tokens initialized with random weights.
    \item \textbf{Average}: Incremental learning with new tokens initialized with the average weights of all other products.
    \item \textbf{Unknown}: Incremental learning with new tokens initialized with the weights of the previously learned unknown token.
\end{enumerate}

These approaches were designed to evaluate different strategies for incorporating new tokens and their impact on the performance of the incremental learning algorithm, as well as to determine the necessity of incremental learning embeddings.

\begin{figure}
    \centering
    \includegraphics[width=0.85\linewidth]{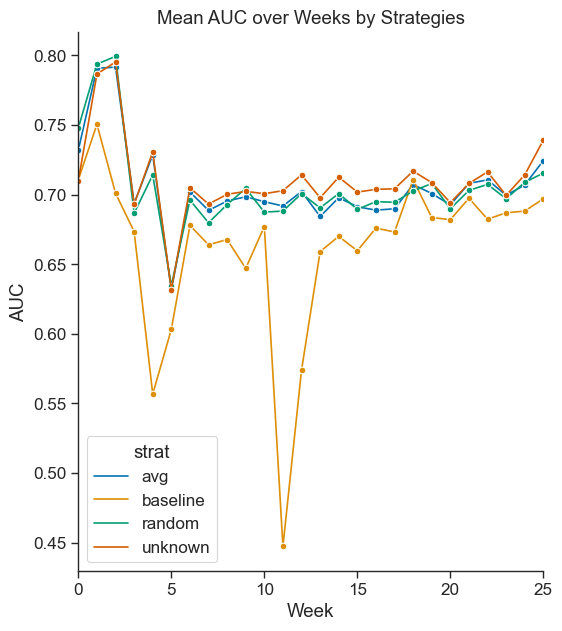}
    \caption{AUC-score of the four tested approaches for each week of the yoochoose dataset.}
    \label{fig:results}
\end{figure}

Figure \ref{fig:results} illustrates the results of our experiments, with performance evaluated using the average AUC score across ten different training runs. The findings consistently demonstrate that incremental learning outperforms the baseline approach across all weeks. Furthermore, the experiments reveal that initializing new tokens with random weights yields poorer results compared to averaging over all tokens. Notably, initializing new tokens with the weights from the unknown token achieves the highest overall performance. The average AUC score and standard deviation over all 26 weeks are $0.662 \pm 0.093$ for the baseline, $0.704 \pm 0.033$ for random, $0.705 \pm 0.032$ for average, and $0.710 \pm 0.032$ for unknown. These results provide preliminary evidence for the effectiveness of incremental learning embeddings, particularly when utilizing the unknown token strategy for initializing new tokens.

\section{Required Future Work for Lifelong-Learning Embeddings}
The approach presented in this paper represents an initial step towards developing lifelong-learning embeddings, allowing for the extension of embeddings by increasing the input size. Our preliminary experiments provide compelling evidence that incremental learning is a valuable method for extending embeddings in e-commerce using our proposed algorithm. However, several critical questions remain to be addressed in future research. Primarily, it is crucial to further evaluate strategies for determining new weights for newly inserted tokens. Consequently, the next necessary step involves conducting more extensive experiments across various use cases and datasets. This includes exploring different learning models and implementing more advanced strategies as previously mentioned. By investigating these aspects, we aim to refine and validate the effectiveness of incremental learning embeddings in diverse e-commerce applications.

Moreover, an essential aspect to consider is the incremental learning of new information without forgetting previously acquired knowledge. This challenge can be effectively addressed by leveraging methodologies from transfer learning and continuous learning~\cite{Sylvestre-AlviseRebuffi.2017,DavidLopez.2017}. These approaches enable the embedding model to accumulate and integrate new knowledge while retaining valuable insights gained from past data and experiences. By employing transfer learning techniques, where knowledge learned from one task is applied to another related task, and continuous learning strategies, which adaptively update the model over time, we can enhance the robustness and adaptability of lifelong-learning embeddings in dynamic e-commerce environments. This approach not only ensures the preservation of valuable knowledge but also facilitates continual improvement and adaptation as new data and insights emerge.

Another consideration is identifying when the dimensionality of the embedding needs to be increased due to the introduction of numerous new tokens, potentially overwhelming the existing vector space and impairing its ability to adequately represent all tokens. From this, several key questions emerge that we aim to address in future work:
\begin{enumerate}
\item At what point does the learned embedding representation become insufficient for an e-commerce use case, and how can this point be measured?
\item How can new knowledge be learned without forgetting past and relevant knowledge using transfer learning and continuous learning?
\item When can knowledge be forgotten because it is no longer relevant, and how can this be determined?
\item At what point does the dimensionality of the embedding become insufficient to capture the diverse information of the input?
\end{enumerate}
Addressing these questions will be essential for advancing towards robust, lifelong-learning embeddings.

\bibliographystyle{ACM-Reference-Format}
\bibliography{01.main}

\end{document}